\documentclass[journal]{IEEEtran}
\usepackage{subfigure}
\usepackage{multirow}
\usepackage{amssymb}
\usepackage{color}

\usepackage{ifpdf}

\usepackage{cite}

\ifCLASSINFOpdf
  \usepackage[pdftex]{graphicx}
  \graphicspath{{../pdf/}}
  \DeclareGraphicsExtensions{.pdf,.jpeg,.png}
\else
  \usepackage[dvips]{graphicx}
\fi
\usepackage{amsmath}
\begin{document}
%
\title{SAIS: Single-stage Anchor-free
Instance Segmentation}
%
%
%

\author{Canqun~Xiang,~
        Shishun~Tian,~
        Wenbin~Zou~and~Chen~Xu
\thanks{C. Xiang,~S. Tian and W. Zou are with Shenzhen Key Laboratory of Advanced Machine Learning and Applications, Guangdong Laboratory of Artificial Intelligence and Digital Economy(SZ), Guangdong Key Laboratory of Intelligent Information Processing, College of Electronics and Information Engineering, Shenzhen University, Shenzhen 518060, China (e-mail: xiangcanqun2018@email.szu.edu.cn).}
\thanks{C. Xu is with College of Mathematics and Statistics, Shenzhen University, Shenzhen 518060, China.}}

%
%

\markboth{Journal of \LaTeX\ Class Files,~Vol.~14, No.~8, August~2015}%
{Shell \MakeLowercase{\textit{et al.}}: Bare Demo of IEEEtran.cls for IEEE Journals}
%



\maketitle

\begin{abstract}
In this paper, we propose a simple yet efficient instance segmentation approach based on the single-stage anchor-free detector, termed SAIS.
In our approach, the instance segmentation task consists of two parallel subtasks which respectively predict the mask coefficients and the mask prototypes.
Then, instance masks are generated by linearly combining the prototypes with the mask coefficients.
To enhance the quality of instance mask, the information from regression and classification is fused to predict the mask coefficients. 
In addition, center-aware target is designed to preserve the center coordination of each instance, which achieves a stable improvement in instance segmentation. 
The experiment on MS COCO shows that SAIS achieves the performance of the exiting state-of-the-art single-stage methods with a much less memory footprint.
\end{abstract}

\IEEEpeerreviewmaketitle

\section{Introduction}
\IEEEPARstart{I}{nstance} segmentation is one of the general but challenging tasks in computer vision.
In generally, instance segmentation can be split into two steps:  object detection, and pixel classification.
So the current instance segmentation task is directly based on advances in object detection like SSD \cite{SSD}, Faster R-CNN \cite{Fasterrcnn}, and R-FCN \cite{RFCN}.
According to the different types of detection architecture, instance segmentation tasks can be divided into two categories, single-stage instance segmentation and two-stage instance segmentation.

The commonly used two-stage instance segmentation methods focus primarily on the performance over speed. 
Due to the using of a cascade strategy, these methods are usually time-consuming. 
In addition, their dependence on the feature localization makes them difficult to accelerate.
Some of the recently proposed one stage instance segmentation methods, eg. YOLACT \cite{YOLACT}, partly solve those problems by dividing the instance segmentation task into two parallel subtasks: prototype mask generation and per-instance mask coefficients prediction.
It is a effective way to speed up existing two-stage methods like Mask R-CNN \cite{Maskrcnn}.
However, in order to represent different shape instances in an image, all those methods above require lots of anchors and memory.

\begin{figure*}
	\centering
	\includegraphics[width=.9\linewidth]{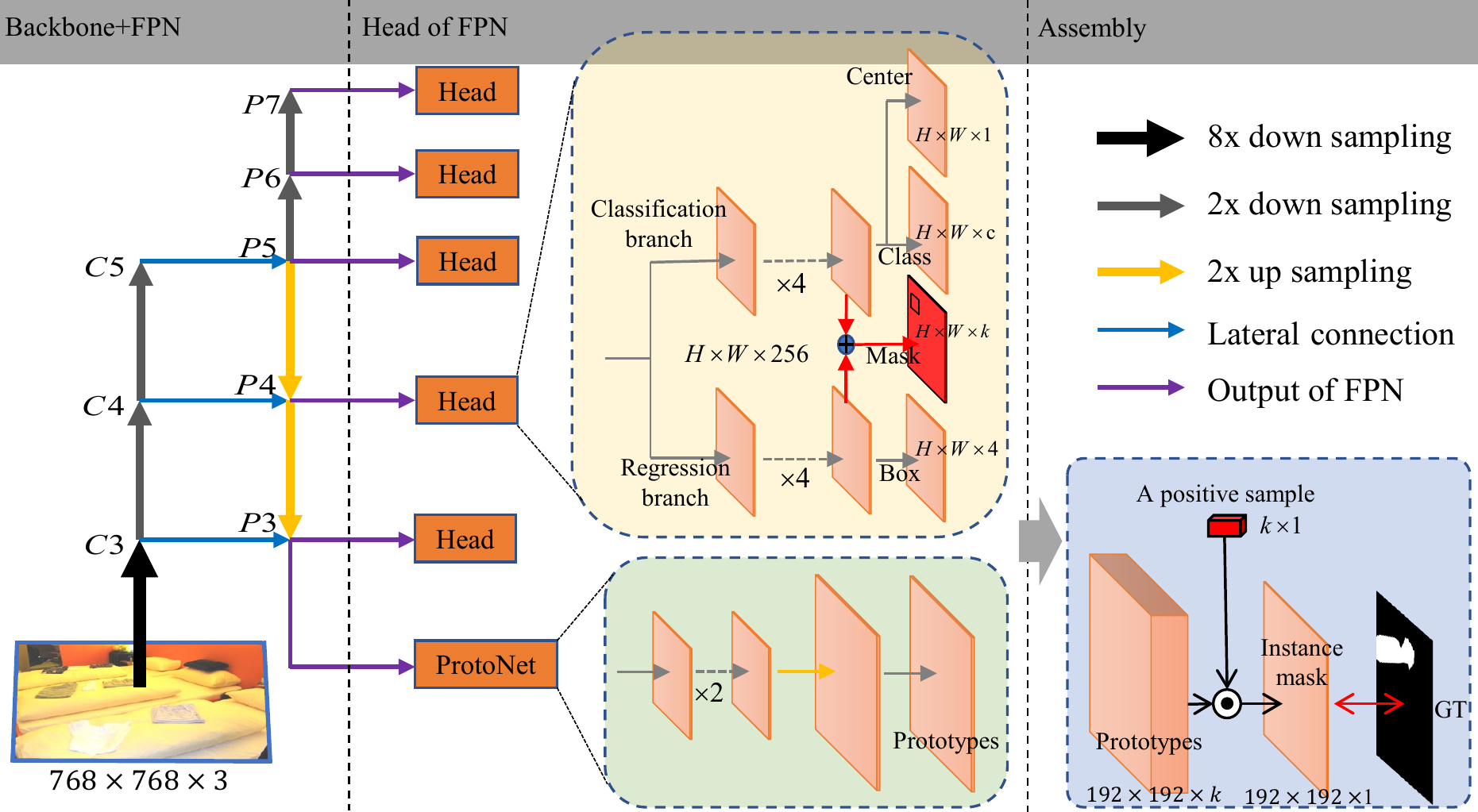}
	\caption{The network architecture of our proposed method, where $C3$, $C4$, and $C5$ denote the feature maps of the backbone network and $P3$ to $P7$ are the feature levels used for the final prediction.
		$H\times{W}$ is the height and width of feature maps.
		In the ProtoNet, Arrows indicate $3\times{3}$ $conv$ layers, except for the final $conv$ which is $1\times{1}$.
		Different lines mean the down-sampling ratio of the level of feature maps to the input image.}
	\label{fig:pipeline}
\end{figure*}

To handle this issue, we propose an instance segmentation method based on the one-stage anchor-free detection framework.
Inspired by some efficient anchor-free detection methods such as FCOS \cite{FCOS}, CenterNet \cite{Centernet1,Centernet2}, etc, which obtain reasonable trade-off between speed and performance by eliminating the predefined set of anchor boxes. 
Based on FCOS, the proposed instance segmentation task is divided into two subtasks similar to YOLACT.
As shown in Fig.\ref{fig:pipeline} (yellow box), one subtask which predicts mask coefficients is assembled into each head of the detector by combining the classification and regression branches.
Only one group of mask coefficients of each sample needs to be predicted since the anchor-free mechanism reduces the total training memory footprint.
The other subtask which generates the prototype masks is directly implemented as an FCN (green box).
All those tasks are implemented in parallel based on single-stage architecture to speed up the training phase. 
Also, to enhance the performance without any additional hyperparameters, we propose a center-aware ground truth scheme, which can effectively preserve the center of each instance during the training and achieve a stable improvement.

Our contributions can be summarized as follows: 
(1) We propose an instance segmentation method based on anchor-free mechanism, which has great advantages in speed and memory usage. 
(2) We propose a center aware ground truth scheme, which effectively improves the performance of our framework in detection and instance segmentation tasks. 
\section{Related work}
\textbf{Two-Stage Instance Segmentation.}
Instance segmentation can be solved by bounding box detection then semantic segmentation within each box, which is adopted by most of existing two-stage methods.
Based on Faster R-CNN \cite{Fasterrcnn}, Mask R-CNN \cite{Maskrcnn} simply adds an mask branch to predict mask of each instance.
Mask Scoring R-CNN \cite{Masksrcnn} re-scores the confidence of mask from classification score by adding a mask-IoU branch, which makes the network to predict the IoU of mask and ground-truth.
FCIS \cite{FCIS} predicts a set of position-sensitive output channels which simultaneously address object classes, boxes, and masks. 
The above state-of-the-art methods can achieve satisfy performance but are time-consuming.

\textbf{Single-Stage Instance Segmentation.}
SPRNet \cite{SPRNet} has an encoder-decoder structure, in which classification, regression and mask branches are processed in parallel. 
It generates each instance mask from a single pixel, and then resize the mask to fit the corresponding box to get the final instance-level prediction. 
In the decoding part, each pixel is used as an instance carrier to generate the instance mask, on which consecutive deconvolutions are applied to get the final predictions. 
YOLACT \cite{YOLACT} divide the instance segmentation into two parallel subtasks, the prototype mask generation and the pre-instance mask coefficient prediction.
Then, the generated prototypes are combined linearly by using the corresponding predicted coefficients and cropped with a predicting bounding box.
TensorMask \cite{Tensormask} investigates the paradigms of dense sliding window instance segmentation by using structured 4D tensors to represent masks over a spatial domain.
All of above methods use anchor-based detection backbone, which requires plenty of memory footprint in the training phase.
Polarmask \cite{Polarmask} formulates the instance segmentation problem as instance center classification and dense distance regression in a polar coordinate.
ExtremeNet \cite{Extremenet} uses keypoint detection to predict 8 extreme points of one instance and generates an octagon mask, which achieves relatively reasonable object mask prediction.
It is a anchor-free method, but the octagon mask encoded method might not depict the mask precisely.
We propose a novel instance segmentation method by combining the single-stage anchor free framework and robust mask encoding method.

\section{Method}
In this section, the proposed method is introduced in detail.
The pipeline is shown in Fig.\ref{fig:pipeline}.
In the section \ref{sec3a}, we explore the application of the anchor-free mechanism on instance segmentation task. 
In the section \ref{sec3b}, we propose a center-aware ground truth to improve the performance.

\subsection{Single-stage anchor-free instance segmentation}\label{sec3a}
YOLACT \cite{YOLACT} is a real-time instance segmentation method in which instance segmentation can be divided into two parallel subtasks: 1) mask coefficients prediction and 2) prototypes prediction.
In this paper, we follow this parallel mechanism to accelerate the model.
\paragraph{Anchor-free for mask coefficients}
Instance segmentation depends strongly on the accuracy of bounding box.
To obtain a high-quality bounding box of an instance, the proposed SAIS is based on the FCOS \cite{FCOS}, an one-stage anchor-free architecture that achieves state-of-the-art performance on object detection tasks.
As shown in Fig.\ref{fig:pipeline},  each head has two branches, one is used to detect 4 bounding boxes regressions,  the other is used to predict 1 center possibility and $c$ class confidences.
Different from FCOS \cite{FCOS}, in each head, a new output layer is added to predict mask coefficients for each sample (Fig. \ref{fig:pipeline} yellow box).
To extract enrich semantic information, we firstly fuse the two branches (classification branch and regression branch) before predicting mask coefficients, followed by a convolutional layer with $k$ channels to predict $k$ mask coefficients of each sample.
In the proposed method, each sample only has $c+1+4+k$ outputs, which has $a\times$ fewer network output variables than the commonly used anchor-based methods with $a$ anchor boxes per sample.

\paragraph{Mask prediction}
Note that the prototype generation branch (protonet) predicts a set of $k$ prototype masks for the entire image.
The protonet is implemented as an FCN whose last layer is with the same channels as the mask coefficient prediction layer.
The final instance masks are generated by combining the mask prototypes and mask coefficients. 
For each sample, the mask coefficient $C$ is produced by the heads of FPN while the mask prototype $P$ is generated by protonet and shared by all samples.
As shown in Fig.\ref{fig:pipeline} (blue box), the final instance mask of this sample is obtained by a single matrix multiplication and sigmoid:
\begin{equation}\label{equ1}
M = \sigma(PC)
\end{equation}
where $C$ is a $k\times1$ matrix and $P$ is an $h\times{w}\times{k}$ matrix. 
The single-stage architecture is composed of the fully convolutional layers, and all subtasks are executed in parallel, which achieves a high speed.

\subsection{Center-aware ground truth}\label{sec3b}
\begin{figure}
	\centering
	\includegraphics[width=0.8\linewidth]{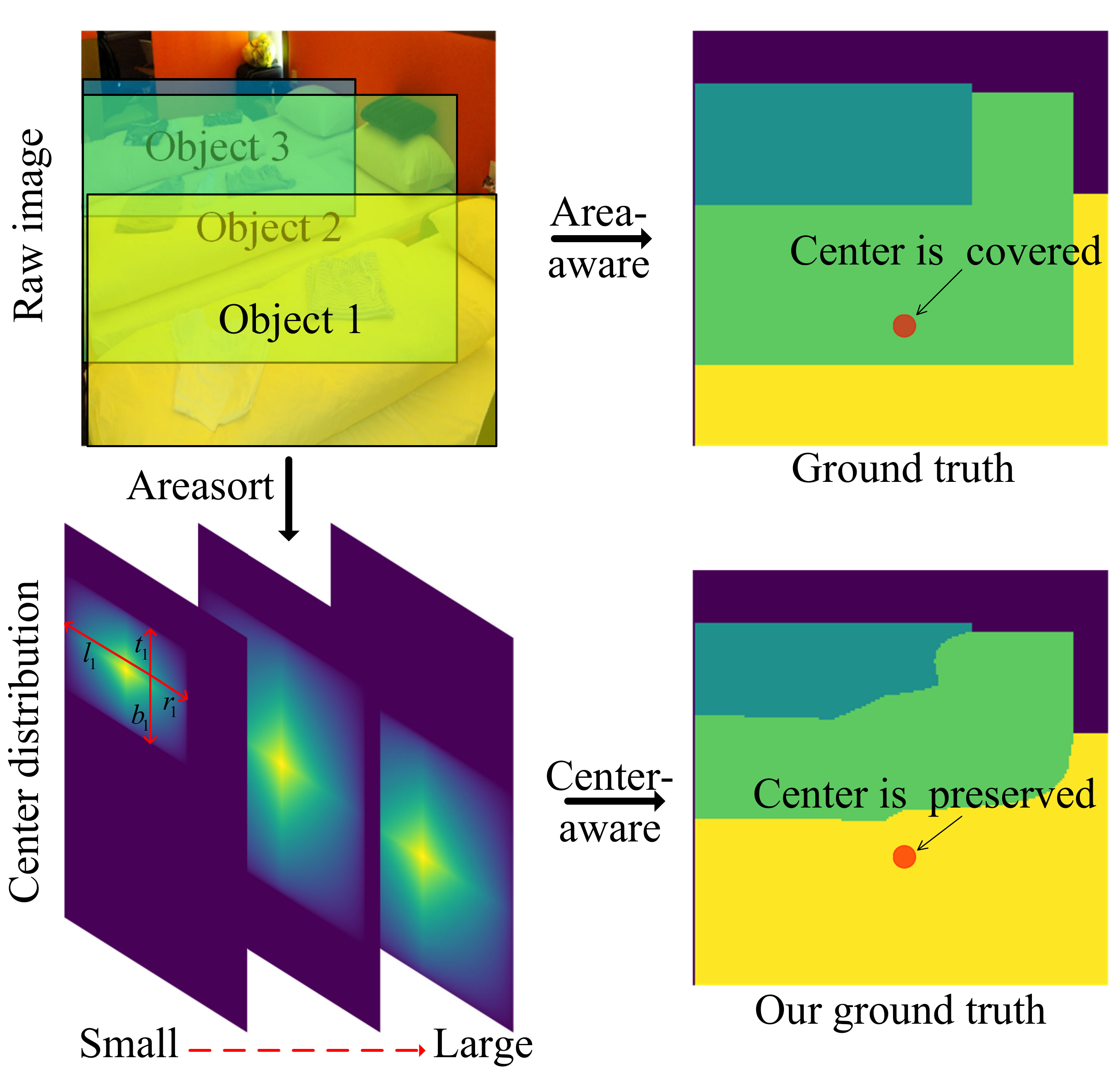}
	\caption{The difference between center-aware ground truth and area-aware ground truth. 
		In those ground truth, each location includes 4 properties (classes, center score, bounding boxes, and instance masks), different colors represent the label from different objects, black means negative samples.}
	\label{fig:center-aware}
\end{figure}

The labels of all tasks are selected from the ground-truth map. 
If a location falls into multiple bounding boxes, it is considered as an ambiguous sample.
To enough consider small objects, a simple way is to choose the bounding box with the minimal area as its regression label as shown in Fig.\ref{fig:center-aware} (top-right).
One big issue is that the center of some large objects may be covered by small objects if the centers of two objects close enough.
It may result in incorrect labels which are selected near the area that the real center is covered by another small object.
As shown in Fig.\ref{fig:center-aware} (red circle in the top-right one), the area in the red circle is the center of object 1, but we select the labels from object 2 as its ground-truth.

To address this issue, we propose a new center-aware method to select reasonable labels. 
In our approach, the center distribution of an object is considered as prior information and makes sure that the real center of each object is preserved in training.
Then we choose the bounding box with the minimal area as its regression label.
Our method can be formally described as follows:
\begin{equation}\label{equ2}
Ind = areasort(a_1, a_2, ..., a_n)
\end{equation}
\begin{equation}\label{equ3}
C_i = \sqrt{\frac{min(l_{Ind(i)}, r_{Ind(i)})}{max(l_{Ind(i)}, r_{Ind(i)})}\times{\frac{min(t_{Ind(i)}, b_{Ind(i)})}{max(t_{Ind(i)}, b_{Ind(i)})}}}
\end{equation}
\begin{equation}\label{equ4}
GT = max(C_1, C_2, ..., C_n)
\end{equation}
where we have $n$ instances in a raw image, $a_i$ means the area of the bounding box of $i$-th instance.
$areasort()$ means all instances are sorted by the size of areas from small to large, which makes sure the small objects are firstly considered.
We calculate the center distribution of each object by Equation (\ref{equ3}), where $l_i$, $r_i$, $t_i$, and $b_i$ are the distances from the location $i$ to the four sides of the bounding box, as shown in Fig.\ref{fig:center-aware} (bottom-left). 
Finally, as shown in Fig.\ref{fig:center-aware} (bottom-right), we choose the object corresponding to the largest score center as the ground truth for each location. 
The area and center distribution are considered in our method simultaneously in the proposed method to achieve better performance.
\begin{table}
	\caption{Comparing the performance of inputs with different types.}
    \centering
	\begin{tabular}{c|c|cc}
		input                   & size          & $mAP^{bbox}$       & $mAP^{mask}$ \\ 
		\hline 
		keeping aspect ratio              & (1333, 800)   & \textbf{36.7}               & 21.2       \\ 
	    fixed size               & (768, 768)    & 35.9               & \textbf{28.2}       \\ 
	\end{tabular} \label{tab0}
\end{table}
\begin{table}
	\caption{Comparing the results on object detection and instance segmentation. 
	'w/ c' means the target with center-aware. For object detection, evaluate annotation type is bbox. For instance segmentation, evaluate annotation type is mask.
    All methods make use of ResNet-50-FPN as backbone}
    \centering
	\begin{tabular}{c|c|ccc|ccc}
		Method                           & w/ c       & $mAP$         & $AP_{50}$     & $AP_{75}$     & $AP_S$        & $AP_M$        & $AP_L$        \\ 
		\hline 
		\multirow{2}{*}{FCOS\cite{FCOS}} & o          & 36.7          & 55.5          & 39.3          & \textbf{21.9} & 40.5          & \textbf{48.0} \\ 
		
		~                                & \checkmark & \textbf{36.9} & \textbf{55.7} & \textbf{39.5} & 21.5          & \textbf{40.9} & 47.3          \\ 
		\hline 
		\multirow{2}{*}{\textbf{SAIS}}   & o          & 28.2         & 47.8          & 29.0          & 9.3          & 30.6          & 44.9          \\ 
	
		~                                & \checkmark & \textbf{28.5} & \textbf{48.4} & \textbf{29.3} & \textbf{10.2} & \textbf{30.7} & \textbf{45.1} \\ 
	\end{tabular} \label{tab1}
\end{table}
\begin{table}
	\caption{Feature fusion for mask coefficient prediction. 
		Comparison the performance w/ (w/o) summed the classification branch and regression branch. If w/o summation, the mask coefficients are predicted only by the classification branch.}
	\centering
	\begin{tabular}{c|ccc|ccc}
		Fusion    & $mAP$         & $AP_{50}$     & $AP_{75}$     & $AP_S$        & $AP_M$        & $AP_L$        \\ 
		\hline 
		w/o          & 28.5      & 48.4      & 29.3   & 10.2 & 30.7 & \textbf{45.1} \\ 
		
		w/        & \textbf{28.7} & \textbf{49.2}          & \textbf{29.4} & \textbf{10.4} & \textbf{32.2} & 44.5          \\ 
	\end{tabular} \label{tab2}
\end{table}
\begin{table*}[tb]
	\caption{Comparison with state-of-the-art. Instance segmentation mask AP on the COCO \emph{test-dev}. The FPS of our model is reported on TITAN X GPUs Better backbones bring expected gains: deeper networks do better, and ResNeXt improves on ResNet.}
	\centering
	\begin{tabular}{c|c|cc|ccc|ccc|c|c|c}
		Method                            & Backbone    & epochs & aug     & $mAP$           & $AP_{50}$     & $AP_{75}$     & $AP_S$        & $AP_M$        & $AP_L$        & FPS  & Mem(GB) & GPU \\ 
		\hline 
		FCIS                              & ResNet-101-C5 & 12 & o  & 29.5            & 51.5          & 30.2            & 8.0           & 31.0          & 49.7          &  6.6    & -  &Xp\\ 
		
		Mask R-CNN      & ResNet-101-FPN  & 12 & o  & \textbf{37.5}            & \textbf{60.2}          & \textbf{40.0}          & \textbf{19.8}          & \textbf{41.2}          & \textbf{51.4}          &  8.6    & 5.7 &Xp\\ 
		\hline 
		ExtremeNet                        & Hourglass-104   & 100 & \checkmark & 18.9            & 44.5          & 13.7          & 10.4          & 20.4          & 28.3          &  4.1    & - &Xp \\
		YOLACT-550                            & ResNet-50-FPN  & 48 & \checkmark  & 28.2            & 46.6          & 29.2          & 9.2           & 29.3          & \textbf{44.8}          &  \textbf{42.5}    & 3.8 &Xp\\ 

		\textbf{SAIS-640}   & ResNet-50-FPN  & 12 & o &  27.6          & 47.2          &  28.3         & 9.4          & 30.5          & 44.0          &  29.2    & \textbf{1.8} &X\\ 
		\textbf{SAIS-768}   & ResNet-50-FPN  & 12 & o &  \textbf{28.7}          & \textbf{49.2}          &  \textbf{29.4}         & \textbf{10.4}          & \textbf{32.2}          & 44.5          &  26.9    & 2.5 &X\\ 	
		\hline
		YOLACT-700                            & ResNet-101-FPN  & 48 & \checkmark  & 31.2            & 50.6          & 32.8          & 12.1           & 33.3      & 47.1          &  23.6    & - &Xp\\  
		\textbf{SAIS-768}                                & ResNet-101-FPN  & 12 & o & 30.7            & 51.6          & 31.7          & 11.3          & 34.3          & 46.8          &  \textbf{25.4}    & \textbf{3.6} & X\\ 
		\textbf{SAIS-768}                                & ResNeXt-101-FPN & 12 & o &   \textbf{32.5}          & \textbf{55.8}          & \textbf{33.6}         & \textbf{13.8}          & \textbf{35.5}          & \textbf{50.3}          &  18.2    & 6.6 &X\\ 
	\end{tabular} \label{tab3}
\end{table*}
\section{Experiments}
We report results on MS COCO's instance segmentation task \cite{MSCOCO} using the standard metrics for the task.
We train on \emph{train2017} and evaluate on \emph{val2017} and \emph{test-dev}.
We implement our method on mmdetection \cite{MMdet}.

\textbf{Training details.}
In our experiments,  our network is trained using stochastic gradient descent (SGD) for 12 epochs with a mini-batch of 16 images.
The initial learning rate and momentum are 0.01 and 0.9 respectively.
The learning rate is reduced by a factor of 10 at epoch 8, 11 respectively.
Specifically, the input image is resized to $768\times768$.
The output channel of protonet is set to 32.
We initialize backbone networks with the weights pretrained on ImageNet \cite{ImageNet}.

\subsection{Ablation study}

\textbf{Fixed Input Size.}
As shown in TABLE \ref{tab0}, we find an interesting phenomenon that fixing the input size achieves a gain of 7\% in term of mask prediction accuracy in comparison with keeping the aspect ratio, even if the size of the former is lower than the latter.
We argue that the inputs with the fixed size make the model easily represent instance-level semantic context. 

\textbf{Center Awareness.}
To evaluate the effectiveness of our proposed center-aware target, we implement our method on two different tasks, object detection and instance segmentation.
FCOS \cite{FCOS} is the state-of-the-art method used for object detection in which the offsets of the bounding box are predicted based on the center position.
The results, shown in TABLE \ref{tab1}, reveal that the center-aware target achieves a gain of 0.2\% and 0.3\% in term of $mAP$ on object detection and instance segmentation tasks respectively. 
We argue that it is important for instance segmentation to predict the masks from the center of object.

\textbf{Feature Fusion.}
To achieve competitive performance, we fuse the feature maps from classification and regression branches to predict the mask coefficients without additional parameters.
The results shown in TABLE \ref{tab2} reveal that the performance gain benefits from the fusion of the feature maps, especially in small instance. 
It is reasonable that bounding box (regression branch) contributes extra information for mask coefficients prediction.

\subsection{Comparison with the state-of-the-art.}
\begin{figure}
	\centering
	\includegraphics[width=1.0\linewidth]{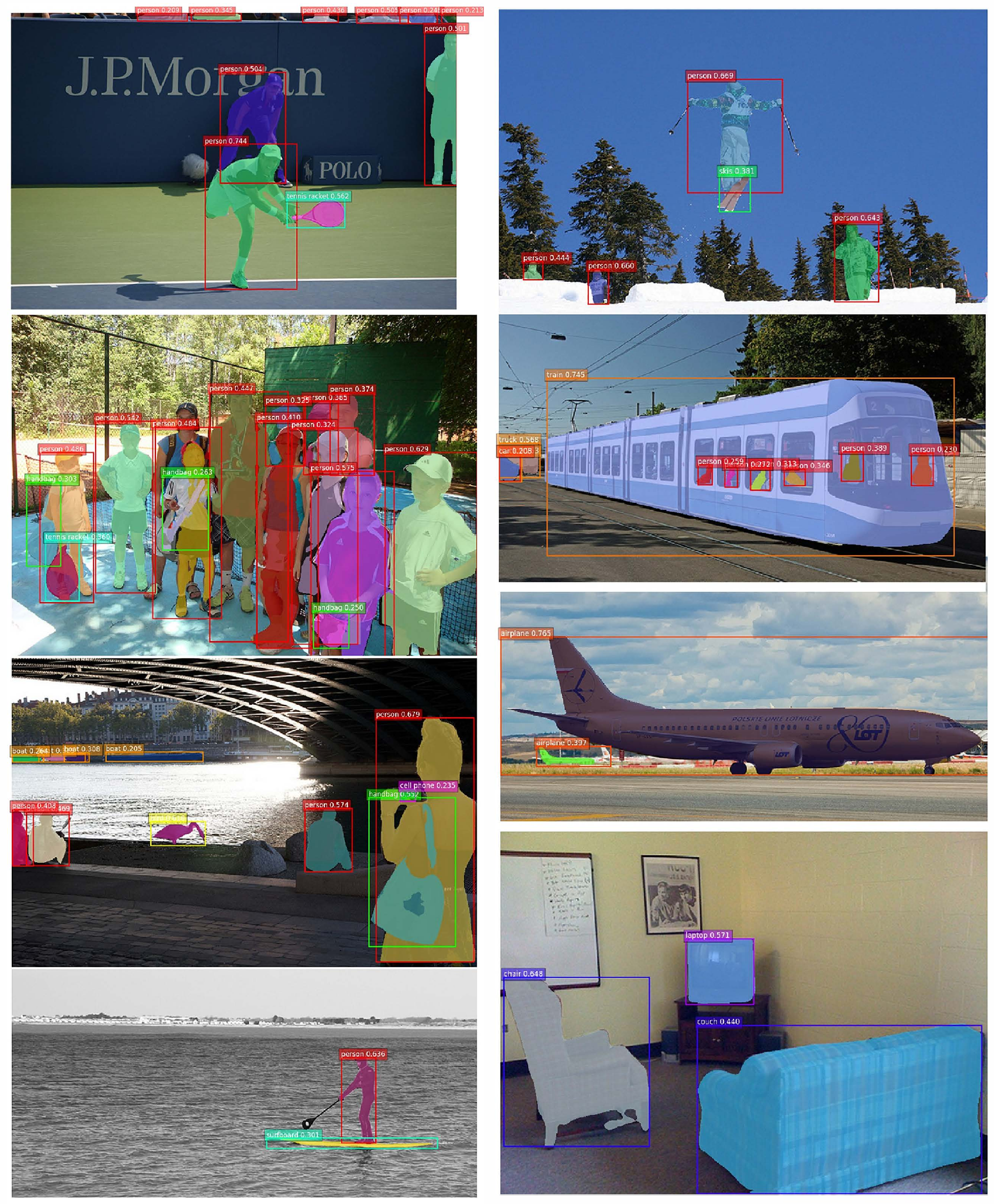}
	\caption{Quantitative examples on the MS COCO \emph{test-dev}. For each image, one color corresponds to one instance in that image.}
	\label{fig:quantityresults}
\end{figure}

In this part, we compare the performance of the proposed method with various state-of-the-art methods including both two-stage and single-stage models on MS COCO dataset.
Our method outputs are visualized in Fig. \ref{fig:quantityresults}.

The results show that, without bells and whistles, our proposed method is able to achieve competitive performance in comparison with one-stage methods.
In less than quarter training epochs without data augmentation and additional semantic loss \cite{YOLACT}, SAIS-768 outperforms YOLACT-550 with the same ResNet-50-FPN backbone and ExtremeNet with Hourglass-104 backbone by 0.5\% and 9.3\% in $mAP$, respectively.
Anchor-free architecture is used in SAIS, which achieves $2\times$ less training memory footprint than all those anchor-based methods.
SAIS-640 with ResNet-50-FPN also achieves 29.2 FPS on TITIAN X GPU without Fast NMS \cite{YOLACT} and light-weight head \cite{YOLACT} that are exploited in YOLACT.
Specially,  SAIS-768 achieves 25.4 FPS over YOLACT-700 with the same ResNet-101-FPN backbone. 
It reveals that anchor-free mechanism is superior to anchor-base in terms of speed.
Compared to two-stage methods, SAIS-640 achieves $3\times$ higher FPS and $3\times$ less memory footprint in the training phase.
In summary, the proposed method, which fuses anchor-free framework and parallel instance segmentation subtasks, achieves competitive performance on speed and accuracy.

The quantity results shown in Fig. \ref{fig:quantityresults} reveal that the quality masks are generated in our method by robust mask encoding method without repooling operation (RoI Pooling/Align \cite{Fasterrcnn,Maskrcnn}) for original feature. 

\section{Conclusion}
In this paper, we propose a single-stage anchor-free instance segmentation method in which all tasks are parallel implemented.
To enhance the performance, a center-aware ground truth is designed without any additional parameters.
Our framework achieves competitive performance on MS COCO dataset.
In the future, we will focus on lightweight framework for instance segmentation, which is a promising direction for industrial applications.

\end{document}